\documentclass[nohyperref]{article}

\usepackage{microtype}
\usepackage{amsmath}
\usepackage{breqn}
\usepackage{graphicx}
\usepackage{subfigure}
\usepackage{booktabs} 

\usepackage{hyperref}

\usepackage[accepted]{icml2022}

\usepackage{amsmath}
\usepackage{amssymb}
\usepackage{mathtools}
\usepackage{amsthm}

\usepackage[capitalize,noabbrev]{cleveref}

\theoremstyle{plain}

\theoremstyle{definition}

\theoremstyle{remark}

\usepackage[textsize=tiny]{todonotes}

\icmltitlerunning{Generative Thermal Design Through Boundary Representation and Multi-Agent Cooperative Environment}

\begin{document}

\twocolumn[
\icmltitle{Generative Thermal Design Through Boundary Representation and Multi-Agent Cooperative Environment}




\begin{icmlauthorlist}
\icmlauthor{Hadi Keramati}{sch}
\icmlauthor{Feridun Hamdullahpur}{sch}


\end{icmlauthorlist}

\icmlaffiliation{sch}{Department of Mechanical and Mechatronics Engineering, University of Waterloo, Waterloo N2L 3G1, Canada}

\icmlcorrespondingauthor{Hadi Keramati}{hkeramati@uwaterloo.ca}

\icmlkeywords{Machine Learning, ICML}

\vskip 0.3in
]



\printAffiliationsAndNotice{} 

\begin{abstract}
Generative design has been growing across the design community as a viable method for design space exploration. Thermal design is more complex than mechanical or aerodynamic design because of the additional convection-diffusion equation and its pertinent boundary interaction. We present a generative thermal design using cooperative multi-agent deep reinforcement learning  and continuous geometric representation of the fluid and solid domain. The proposed framework consists of a pre-trained neural network surrogate model as an environment to predict heat transfer and pressure drop of the generated geometries. The design space is parameterized by composite Bézier curve to solve multiple fin shape optimization. We show that our multi-agent framework can learn the policy for design strategy using multi-objective reward without the need for shape derivation or differentiable objective function.  
\end{abstract}

\section{Introduction}
\label{Introduction}

Additive Manufacturing allows conversion of complex digital geometries to functional real world objects \cite{keramati2022deep}. Therefore, Generative Design (GD) receives more attention as a method to generate optimized geometry. Generative design is a design exploration process performed iteratively over various geometries followed by a performance evaluation \cite{oh2019deep}. There are several GD techniques introduced recently including Generative Adversarial Networks (GANs), Variational Autoencoders (VAEs), and deep Reinforcement Learning (RL) \cite{regenwetter2022deep,viquerat2020supervised}. The latter is different from the first two methods since RL works as a semi-supervised Machine Learning (ML) algorithm.

GANs generate new designs from an existing dataset utilizing a generator and a discriminator which are usually Deep Neural Networks ( DNNs ). The objective function of GANs should be differentiable to utilize gradient-based optimization while reward of a deep RL can be defined based on the design requirements \cite{chen2021padgan}.

Shape and Topology Optimization ( TO ) play a major role in Generative models in engineering design \cite{chen2021mo}. Engineering design often require Finite Element Analysis (FEA) or Computational Fluid Dynamics (CFD) to assess the performance of the output design \cite{hoyer2019neural}. These numerical approaches are computationally expensive and require human expertise \cite{regenwetter2022deep}. Design of heat transfer devices is more challenging than structural or fluid flow due to higher order of nonlinearity in the numerical simulation \cite{feppon2020topology}. Gaussian process regression also known as Kriging metamodel was used as a geostatistical estimator of the CFD computation to accelerate design automation  \cite{zadeh2019efficient, liu2021trust}. In this work, we utilize a surrogate model based on Convolutional Neural Network (CNN) to estimate the CFD results directly from geometries which is then used as a computational engine for RL environment. Baque et al. \cite{baque_geodesic_2018} used CNNS as a surrogate model for CFD of aerodynamic shapes along with gradient-based optimization in which the objective function should be differentiable with respect to the shape parameters. Mark C. Messner assessed the accuracy of a CNN surrogate model for predicting mechanical properties which was then used for topology optimization \cite{messner2020convolutional}.

Several methods have been used for topology optimization such as density based, phase field, and shape derivative methods \cite{feppon2020topology}. The density-based topology optimization has been used for decades in structural optimization; however, this method is not practical for thermo-fluid applications since a porous medium approach is used for solid distribution. In fluid flow and particularly heat exchanger design, boundary conditions are a significant part of the simulation and it is important to have distinguishable boundaries. A naive way of optimizing the design while keeping the boundaries distinguishable is using pixel or voxel-based optimization which require high CPU time due to the curse of dimensionality \cite{mekki2021genetic}. Feppon et al. \cite{feppon2020null} used Hadamard’s boundary variation method along with adjoint-based solver to keep a clear boundary condition throughout the shape optimization.

The agent of a deep RL which is a DNN learns through iterative interaction with the environment \cite{mnih2015human}. Deep RL was successfully implemented for modeling 3D shapes \cite{lin2020modeling} as well as structural TO \cite{brown2021deep} of a Cantilever with elementally discretized domain.

In this study, we use Multi-Agent Reinforcement Learning ( MARL ) for heat exchanger shape optimization in which a surrogate model is used as an environment. Parametric curves are used as the geometry representation which facilitate the implementation of boundary conditions and design space exploration in contrast to traditional topology optimization. We utilize Centralized Training and Decentralized Execution (CTDE) MARL to address the non-convergence in continuous action space.

\section{Background}
Below we provide background on Boundary Representation ( BREP ) and Proximal Policy Optimization ( PPO ).

\subsection{BREP-Based Geometry}

There are several design representation including Signed Distance Functions (SDFs), pixels and voxels, point clouds, mesh, grammars, and BREP. These design representations are often converted to BREP for downstream tasks such as simulation and manufacturing \cite{seff2021vitruvion}. Bézier curve is one of the basic components of BREP and Composite Bézier curve is a piece-wise Bézier curve where the initial and final points are joined together to form a continuous geometry. The key component that makes Bézier curves well-suited for heat transfer interaction is the continuous derivative on the curve which is extremely useful for Neumann boundary condition implementation. 
\noindent
The original Bézier curve with a set of control points $ P_i$ with $n+1$ parameters are defined as following: 
\begin{equation}\label{bspline}
	\alpha(u) = \sum_{i=0}^n P_i B_{i,n}(u) ; u \in [0,1]
\end{equation}
\noindent Where $ B_{i,n}(u)$ is the \textit{i}'th function of degree \textit{n} defined by Bernstein polynomials
\begin{equation}\label{bspline2}
	B_{i,n}(u) = {n \choose i} n^i (1-u)^{n-i} ; i=0, .., n
\end{equation}


\subsection{Proximal Policy Optimization ( PPO )}

In this work, we use PPO \cite{schulman_proximal_2017} as our main RL algorithm. PPO benefits from improved stability of the stochastic policy updates during training particularly in the presence of potential occasional divergence in computation environment. PPO updates policies via Eq. \ref{ppoupdate} \cite{schulman_proximal_2017}.

\begin{equation}\label{ppoupdate}
	\theta_{k+1} = \arg \max_{\theta} \underset{s,a \sim \pi_{\theta_k}}{{\mathrm E}}\left[
	L(s_t,a_t,\theta_k, \theta)\right]
\end{equation}

\noindent where $L$ in PPO-Clip method is a surrogate objective according to the Eq. \ref{lfunc}.
\begin{equation}\label{lfunc}
	\begin{aligned}		
		L(s_t,a_t,\theta_k,\theta) =  \min\left(
		\frac{\pi_{\theta}(a_t|s_t)}{\pi_{\theta_k}(a_t|s_t)}  A^{\pi_{\theta_k}}(s_t,a_t),g \right) \\ 	
	\end{aligned}
\end{equation}
where
\begin{equation}
	g =\text{clip} (\frac{\pi_{\theta}(a_t|s_t)}{\pi_{\theta_k}(a_t|s_t)}, 1 - \epsilon, 1+\epsilon) A^{\pi_{\theta_k}}(s_t,a_t)   
\end{equation}

\noindent in which $ \pi_{\theta} $ denote a policy with parameters $\theta$,  $\epsilon$ is a hyperparameter controlling the correlation between new and old policy considered to be 0.2 in this study, $A^{\pi_{\theta}}$ is the advantage function for the current policy. $a_t \in \mathcal{A}$ and $s_t \in \mathcal{S}$ are action and state from action and state space $\mathcal{A}$ and $\mathcal{S}$. Advantage $A^{\pi_{\theta_k}}(s_t,a_t)$ is a function of value function ( $V(s_t)$ ) and reward (  $r_t$ ).

\section{Methodology}
\begin{figure}[htb]
	\centering
	\subfigure[]{\includegraphics[width=0.225\textwidth]{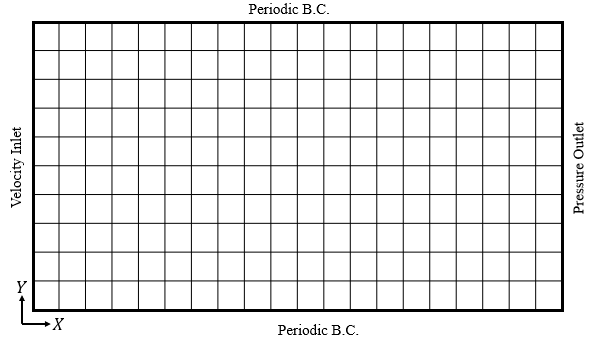}} 
	\subfigure[]{\includegraphics[width=0.24\textwidth]{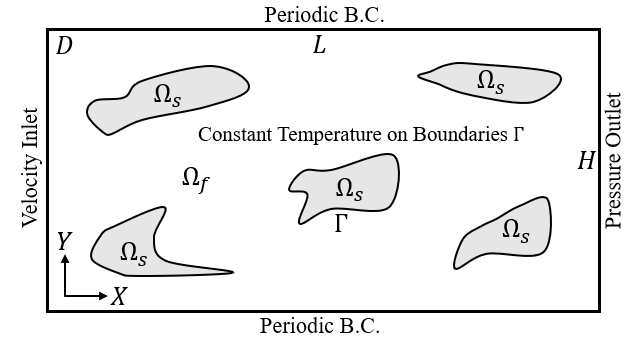}} 
	\caption{Setting of design space exploration for (a) discrete space, and (b) BREP space}
	\label{Design_space}
\end{figure}

The design space is a two dimensional space $ D= \overline{\Omega_f} \cup \overline{\Omega_s} \subset \mathbb{R}^2 $ occupied by solids and an incompressible fluid for Eulerian simulation. The solid components ${\Omega_s}$ consist of composite Bézier controlled by moving control points which causes incremental changes in the fluid domain $\Omega_f$. The BREP design space is shown in Fig. \ref{Design_space} along with the discrete space used in conventional TO.

\subsection{High Fidelity Simulation Environment}

CFD environment is created for the design space shown in Fig. \ref{Design_space} both as a high fidelity simulation and as a surrogate model. The high fidelity environment is written utilizing FEniCS open source software all in python \cite{logg_automated_2012}. FEniCS is based on the Finite Element Method(FEM) used to solve Navier–Stokes and convection-diffusion equations. Boddy-fitted mesh generation of the geometry is performed based on adaptive method and heuristics in GMSH \cite{geuzaine2009gmsh}. 
\noindent Periodic boundary condition is implemented in the vertical direction. Dirichlet boundary for the fluid variables $\textit{\textbf{u}} = \textit{\textbf{u}}_{0}  = 1 m/s$ is considered for the inlet flow as well as Neumann boundary for the fluid variables $\sigma_f (\textit{\textbf{u}},p) \cdot \textit{\textbf{n}} = 0$ at the outlet. Isothermal boundaries for the inlet temperature and solid-fluid interface is considered $ T_{in}  =  300 K$ , $ T_{s}  = 450 K$, respectively.

\noindent Heat transfer and pressure drop are computed between the inlet and outlet boundaries specified in Fig. \ref{Design_space} over the final half of the computation. Final time of the computation is considered to be twice the value of the number of timesteps required for convergence based on the Courant-Friedrichs-Lewy condition.

\subsection{Surrogate Model Environment}
Increasing number of shapes and control points results in larger action space which requires more iteration for policy learning. Therefore, we trained an Xception network \cite{chollet2017xception}  as a surrogate model to predict heat transfer and pressure drop directly from the generated geometries. 34000 geometries generated using 4 and 5 control points along with their CFD results are used for training the surrogate model. The size and resolution of the images are considered to be constant with a single channel, and 506 $\times$ 506 pixels. The pre-trained Xception network predicts heat transfer and pressure drop values with less than 4\% error.

\subsection{Training the Agent}

\noindent The action of an agent is to change the position of the control point presented in Eq. \ref{bspline}. The state of each agent is the position of control points of the pertinent shape. One episode in the environment is defined as the finished simulation of the newly constructed domain based on a set of actions. Parallel training is used in both CFD and surrogate environment with a batch size of 40 and 50. DNN architecture of the deep RL in this study includes three consecutive hidden layer of size 256 and 512 for single-agent RL and MARL, respectively. The classic rectified linear unit (ReLU) function is used as an activation function applied to hidden layers. 

\section{Experiments}

Our experiments are performed under multiple flow and heat transfer conditions controlled by Reynolds ( Re ) and Prandtl number ( Pr ). Reward is the desired output which is heat transfer divided to cost which is pressure drop. An ideal fin is expected to cause maximum heat transfer and minimum pressure. In this study, we consider the reward function to be calculated as Eq. \ref{reward} which an accepted objective in heat transfer community. The agent finds an optimal policy that can maximize the discounted cumulative reward. The reward function, presented in the Eq. \ref{reward} guides the network toward the pressure drop reduction and heat transfer enhancement.
\begin{equation}\label{reward}
	r_t =\frac{Q}{{Dp}^{1/3}}
\end{equation}

\subsection{Multi-Agent Reinforcement Learning}

\begin{figure}[htb]
	\center
	
	\includegraphics[width=0.95\columnwidth]{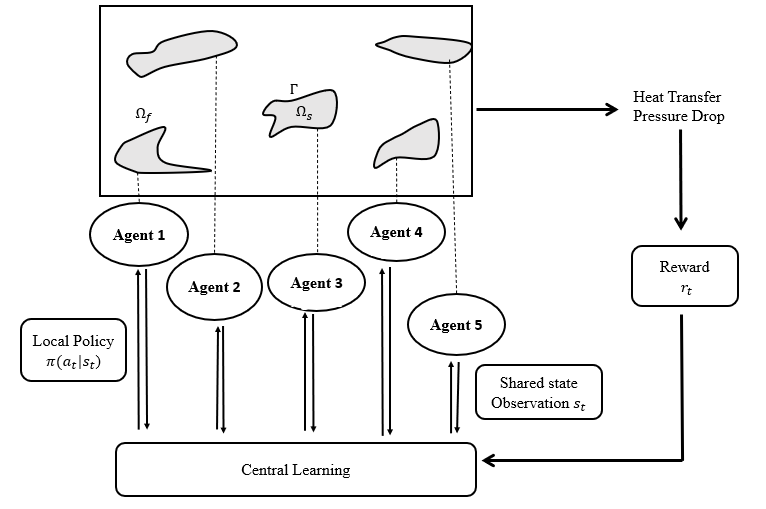}
	\caption{Multi-agent framework for multiple shape optimization }\label{MARL}
\end{figure}

Multi-Agent PPO (MAPPO) shares the same algorithm as the single-agent PPO. IN MAPPO, value function receives global state which reduces variance for policy learning. We follow the Centralized Training and Decentralized Execution (CTDE) with homogeneous agents utilizing parameter sharing. Fig. \ref{MARL} shows our framework.
Each agent is responsible for geometry of one single shape taking action $ a_i$ using shared policy $ \pi_{\theta} $ with parameters $\theta$. The heat transfer domain is a combination of geometries produced by actions $ A = (a_1,...,a_n)$ in which $n$ is the number of agents. Each shape is allowed to occupy a rectangle with the size of $H/4$ and $L/3$ centered by the initial shape. Therefore, all agents act in the same environment with the same action space.

All experiments were conducted on a single workstation equipped with Ubuntu 20.04 LTS, 16-Core Processor 3.40 GHz CPU, 32.0 GB RAM, and one GeForce RTX 3080 Graphics Processing Units (GPUs). Python 3.8.9, TensorFlow, Keras, and Tensorforce are used for ML implementation, as well as Multiprocessing to leverage multiple processors for parallel computing. 

\begin{figure}[htb]
	\centering
	\subfigure[]{\includegraphics[width=3.3cm, height=2.55cm]{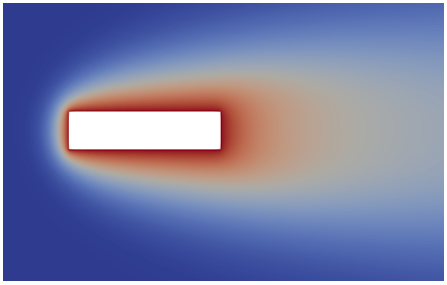}} 
	\subfigure[]{\includegraphics[width=3.9cm, height=2.55cm]{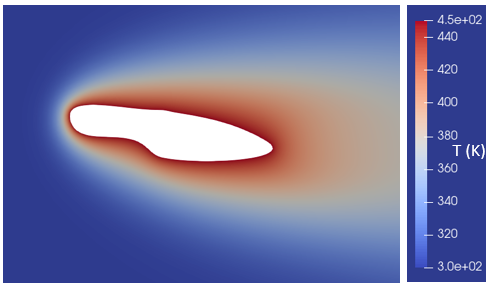}} 
	\caption{Temperature profile for (a) the reference geometry, and (b) optimized geometry in case of single-agent framework}
	\label{optimized}
\end{figure}

\begin{figure}[htb]
	\centering
	\subfigure[]{\includegraphics[width=3.3cm, height=2.53cm]{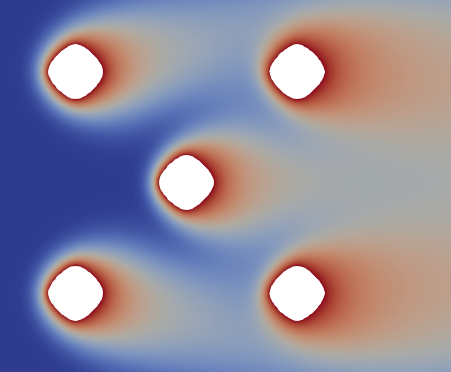}} 
	\subfigure[]{\includegraphics[width=3.95cm, height=2.55cm]{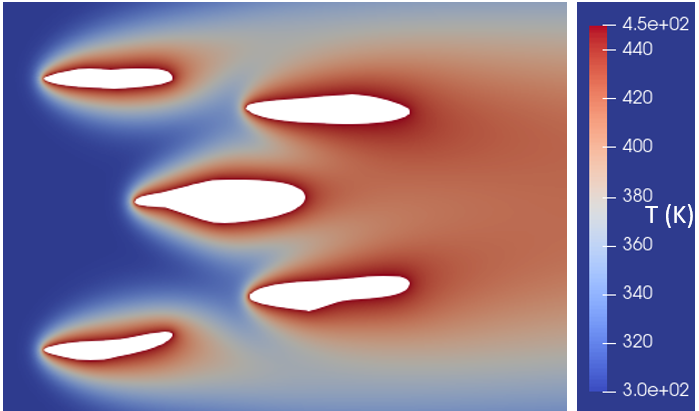}} 
	\caption{Temperature profile for (a) the reference geometry, and (b) one of the well-performing designs in case of multi-agent framework}
	\label{optimized_MARL}
\end{figure}

Fig. \ref{optimized} shows the reference geometry along with one of the well-performing geometries for single-agent framework at Re = 100 and Pr = 0.05. Optimized shapes show over 30 percent improvement in overall heat transfer while lowering the pressure drop by more than 60 percent compared to the rectangle reference geometry. Fig. \ref{optimized_MARL} shows the design for MARL framework at Re = 10, and Pr = 0.7. It can be seen starting from a reference geometry, agents find the policy to optimize the shapes. However, agents fail to establish symmetry in geometry while the physics of the problem is symmetric since each agent acts independently. 

\begin{figure}[htb]
	\centering
	\includegraphics[width=7.4cm]{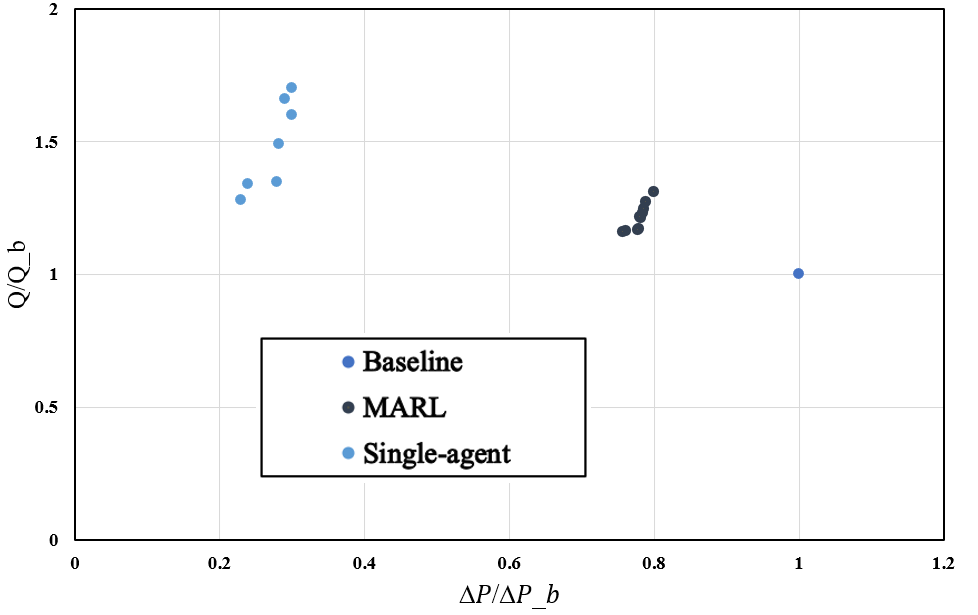}
	\caption{Pareto front of dimensionless heat transfer and pressure drop with respect to the reference geometry for the single-agent and MARL framework }\label{pareto}
\end{figure}

Fig \ref{pareto} shows the dimensionless heat transfer and pressure
drop with respect to the reference geometry for well-performing designs in both frameworks. Heat transfer is affected by fluid flow properties such as Pr and Re number. Lower Re and higher Pr number in MARL case result in lower heat transfer per shape. Multiple shapes also impose higher pressure drop across the domain.

\begin{figure}[htb]
	\centering
	\subfigure[]{\includegraphics[width=7.1cm]{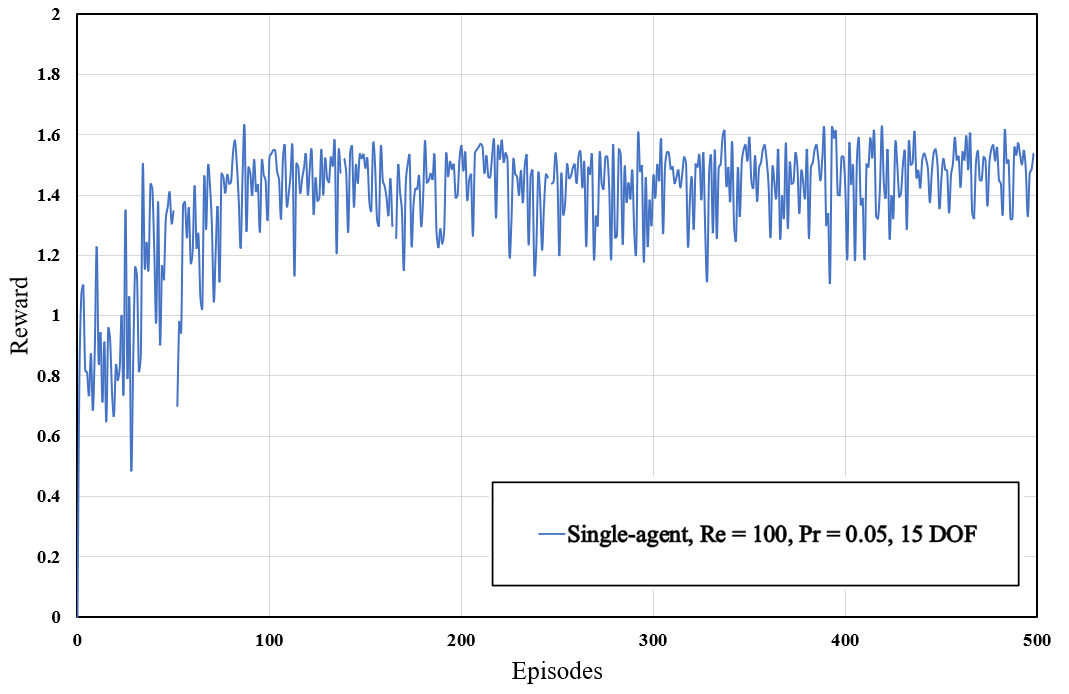}} 
	\subfigure[]{\includegraphics[width=7.1cm]{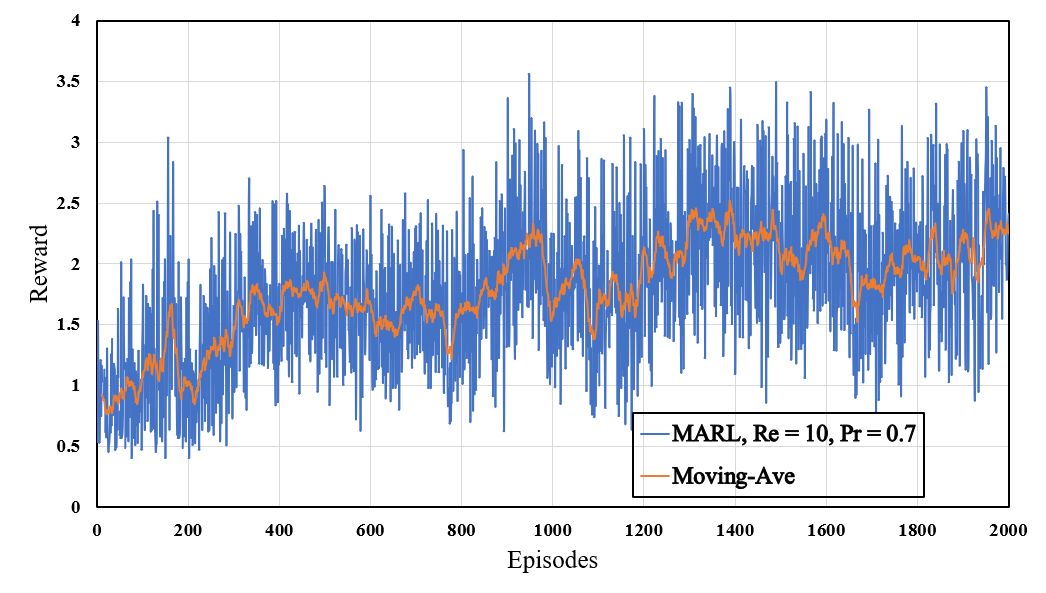}} 
	\caption{Reward history for (a) single agent framework (b) multi-agent framework }
	\label{reward_hist}
\end{figure}

Fig. \ref{reward_hist} shows the reward history during the training of the single agent as well as MARL framework. Reward of the single shape with 15 Degree of Freedom (DOF) reaches a plateau after less than 300 episodes of training. MARL framework with five agents has a large action space. However, after less than 2000 episodes and two hours of training, well-performing design with increased performance are generated.  

\section{Conclusion}
The presented wrok utilized multi-agent deep reinforcement learning to generate multiple optimized fin shapes directly from boundary  representation. A parametric geometry represented by composite Bézier curve is used for design space exploration. We used centralized training and decentralized execution to train reinforcement learning agents for the large continuous action space of up to five control points for each shape. An Xception network is used as a surrogate model to provide accelerated iterative process for agents. Utilizing accelerated experience of the agents from the surrogate model, multiple shapes with enhanced performance are generated.

\newpage
\nocite{langley00}

\bibliography{example_paper}
\bibliographystyle{icml2022}


\newpage
\appendix
\onecolumn

\end{document}